\documentclass[letterpaper, 10 pt, conference]{ieeeconf}  

\IEEEoverridecommandlockouts                              

\usepackage{graphics} 
\usepackage{epsfig} 
\usepackage{mathptmx} 
\usepackage{times} 
\usepackage{amsmath} 
\usepackage{amssymb}  
\usepackage{hyperref}
\usepackage{booktabs}

\title{\LARGE \bf
Estimating Object Physical Properties from 
RGB-D Vision and Depth Robot Sensors Using Deep Learning
}

\author{\authorblockN{Ricardo Pedreiras Cardoso} \authorblockA{Instituto Superior T\'ecnico \\
University of Lisbon\\
Lisboa, Portugal \\
ricardoreisc@tecnico.ulisboa.pt}
\and \authorblockN{Plinio Moreno} \authorblockA{Institute for Systems and Robotics\\
Instituto Superior T\'ecnico\\
University of Lisbon \\
Lisbon, Portugal\\
plinio@isr.tecnico.ulisboa.pt}
}

\begin{document}

\maketitle
\thispagestyle{empty}
\pagestyle{empty}

\begin{abstract}

Inertial mass plays a crucial role in robotic applications such as object grasping, manipulation, and simulation, providing a strong prior for planning and control. Accurately estimating an object's mass before interaction can significantly enhance the performance of various robotic tasks. However, mass estimation using only vision sensors is a relatively underexplored area. This paper proposes a novel approach combining sparse point-cloud data from depth images with RGB images to estimate the mass of objects. We evaluate a range of point-cloud processing architectures, alongside RGB-only methods. To overcome the limited availability of training data, we create a synthetic dataset using ShapeNetSem 3D models, simulating RGBD images via a Kinect camera. This synthetic data is used to train an image generation model for estimating dense depth maps, which we then use to augment an existing dataset of images paired with mass values. Our approach significantly outperforms existing benchmarks across all evaluated metrics. The data generation\footnote{\url{https://github.com/RavineWindteer/ShapenetSem-to-RGBD}} as well as the training of the depth estimator\footnote{\url{https://github.com/RavineWindteer/GLPDepth-Edited}} and the mass estimator\footnote{\url{https://github.com/RavineWindteer/Depth-mass-estimator}} are available online.

\end{abstract}

\section{Introduction}

Planning how to grasp objects benefits from accurately estimating the mass before physical interaction with the object. While mass can be directly measured using a scale, this approach is impractical for robotic manipulation in dynamic or unstructured environments, where physical interaction before grasping is often not feasible. Instead, estimating the mass of an object using visual information from a camera offers significant advantages. A vision-based approach allows robots to predict mass without prior contact, enabling faster and more adaptive decision-making when grasping or lifting objects. Additionally, this method can be more flexible, as it does not require specialized hardware like force sensors or scales.

One major challenge in mass estimation from RGB data alone is the inherent ambiguity in determining object size and mass without sufficient visual context \cite{b18}. Existing datasets attempt to mitigate this by including bounding boxes, and some models predict low-resolution thickness masks to improve estimation accuracy \cite{b1}. However, these approaches are of little use in practice since they require robots first to identify the bounding box of the object they wish to interact with, a time-consuming process often requiring multiple views around the object. Depth sensors can avoid this limitation since they can disambiguate the object's scale from the metric depth. However, previous works rarely made direct use of them, mainly due to the scarcity of datasets that integrate depth information, RGB images and mass. Consequently, no current models fully exploit the combination of depth and image data for mass estimation across a wide range of objects.

To address these limitations, we propose augmenting an existing RGB-mass pairs dataset with depth. We achieve this by first fine-tuning a GLPDepth model \cite{b19} on a synthetic dataset made of RGB and depth pairs, and then applying the depth estimation to the whole dataset. We developed a multimodal architecture that takes as input an RGB image and a sparse point-cloud and outputs a mass prediction in Kg. We propose several variations of the architecture, and we train them on the augmented dataset. We also explore if training the model for point-cloud reconstruction simultaneously improves its mass estimation capabilities. Our results show that our method significantly outperforms previous approaches across all metrics evaluated and that performing point-cloud reconstruction simultaneously does not change the model's predictive power in a significant way.

\section{RELATED WORK}


Mass estimation is often reformulated as two subproblems: (1) estimating volume and (2) estimating density. While volume can be derived through depth cameras or inferred from visual cues and background context, density estimation poses a much greater challenge \cite{b1}. Most existing research focuses on specific, well-constrained problems where mass estimation is simplified by making assumptions about the material properties, effectively reducing the problem to volume estimation. Even so, certain types of objects, such as containers, cans, or boxes, still introduce significant ambiguity. In such cases, mass estimation becomes inherently difficult, as internal contents may be occluded.

Some methods attempt to estimate mass through geometric features like object contours, size, or shape, often by correlating these features with known object properties. For example, in \cite{b2}, the mass of fish was shown to be proportional to their outline, captured via a stereo camera setup. The authors used this knowledge to create a device that accurately captured the mass of swimming fish. This insight led to successful applications in livestock mass estimation, including pigs \cite{b3}, chickens \cite{b4}, and other animals \cite{b5}, as well as fruits \cite{b6}\cite{b7} and industrial products \cite{b8}.

Some have integrated depth data to improve mass estimation, as demonstrated in \cite{b12} and its follow-up \cite{b13}, where a Convolutional Neural Network (CNN) was used to classify and estimate the mass of scrap metals. These models employed both RGB and depth data together and achieved good results. In \cite{b12}, the authors also explored the impact of using a single network for RGB and depth, treating depth as a fourth channel versus having two CNNs, where the network processes RGB images and depth data separately before concatenating them into a single image. Their experiments revealed that the latter approach consistently yielded better results across multiple experiments. However, similarly to the previous approaches, these models were designed for constrained environments with limited object types, which greatly restrained their generalizability to broader scenarios.

A more general approach is to use data-driven methods to infer the volume and mass of the object directly. The most notable example is the \textit{image2mass} model \cite{b9}, which addresses general mass estimation across a diverse set of objects. The authors built a large dataset of approximately 150,000 items scraped from Amazon.com, each annotated with an image, mass, and bounding box. This dataset covers a broad range of categories, including electronics, home goods, and packaged foods, making it more applicable to real-world scenarios compared to earlier, more domain-specific datasets.

The \textit{image2mass} architecture uses two Xception CNNs \cite{b24}, one to predict a coarse 19x19 thickness mask of the object alongside 14 handcrafted features (Geometry Module) and another to estimate the object's density (Density Tower). A third module, called Volume Tower, is used to estimate the volume of the object. The Geometry Module is not trained together with the rest of the model, but rather is used as a sort of preprocessing step. The outputs of this module are then propagated to the two towers. The model makes the assumption that the object in the image has uniform density, which, while not true for most cases, does hold up in empirical experiments. Unlike the previous approaches, this model does not explicitly model volume and density. Instead, the model fuses the outputs of the Density and Volume Towers by multiplying them together. The volume tower focuses on the object's geometric properties, using features extracted from the bounding box and other visual cues, while the density tower aims to infer material properties, effectively learning a distribution of possible densities from the training data. Subsequent research has expanded on this framework, such as in \cite{b10}, which introduced material embeddings into the \textit{image2mass} architecture, and \cite{b11}, which introduced voxel grid estimation for better predictions of the thickness mask. Despite these developments, neither did improve on the original \textit{image2mass} results.

\section{DEPTH PREDICTION DATASET}

Since  no large-scale dataset exists that simultaneously provides RGB images, depth information, and corresponding mass values, we opted to augment the \textit{image2mass} dataset \cite{b9} to include depth, given the substantial overlap between the tasks of mass estimation from RGB and RGB and Depth (RGBD).

To that effect, we fine-tuned a pre-trained GLPDepth model \cite{b14} on synthetic data mimicking the \textit{image2mass} dataset. We built this using the ShapeNetSem \cite{b15} 3D models by first filtering to exclude all those that did not contain valid metric dimensions and weight. This resulted in a total of 8,948 distinct models. Then, we simulated a Kinect camera and captured 14 RGBD images per object, 8 from equidistant rotations at a bird's eye view and 6 for the top, bottom, front, back, left and right.

To ensure all objects appear roughly the same size in the image, we approximate each to a sphere with a diameter equal to its bounding box's diagonal. For every sphere to take up the same amount of space on the screen, the camera's distance must scale with the sphere’s radius. In practice, we position the camera at a distance of 2.1 times this diagonal for consistent object sizing.

By dividing the depth by the diagonal of the object’s bounding box, we normalize the depth, making it scale-agnostic. This allows the model to learn a scale-invariant depth map, which can later be converted back to metric units by multiplying by the bounding box diagonal. This approach is beneficial because the \textit{image2mass} dataset lacks contextual size cues, and it leverages the available bounding box data. After training the GLPDepth model, we apply it to the \textit{image2mass} dataset, creating an augmented dataset with dense depth information for training our model.

\section{MODEL ARCHITECTURE}

\begin{figure*}[thpb]
    \centering
    \setlength{\fboxrule}{0pt}
    \framebox{\includegraphics[width=0.95\textwidth]{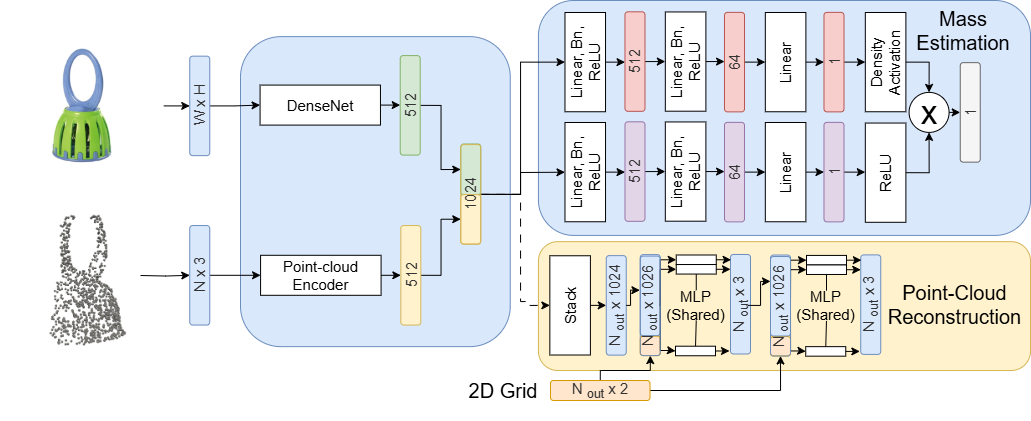}}
    \caption{Our architecture for mass estimation. The leftmost blue block works as an encoder for the point cloud and RGB, and rightmost blue block works as the mass decoder. The yellow block is a FoldingNet architecture, which can be integrated optionally.}
    \label{fig:model_full}
\end{figure*}

Here, we present our model architecture for mass estimation from RGB and Depth, as shown in figure \ref{fig:model_full}. It follows an encoder-decoder architecture. Similar to previous works, we make the assumption that the density of the objects is uniform. In practice, the model learns to estimate the average density of the object. We have two distinct decoders, one for density and another for volume were both take the latent vector generated by the encoders as input. The final mass is predicted by multiplying the outputs of the decoders together.

Several methodologies exist for encoding RGBD data, such as treating them as a single 4-channel image, two distinct images, an image and a point-cloud or by voxelizing the depth data. We argue that treating depth data as a point-cloud rather than an image provides significant advantages. (1) This method enables the model to leverage three-dimensional spatial information more effectively, (2) offers greater adaptability to different sensor intrinsic parameters, and (3) is adaptable to various depth sensors beyond traditional RGB-D cameras, such as LIDARs and SONARs, which produce point-clouds instead of single-image representations. 

Our experiments primarily focus on evaluating how depth processing can enhance the mass estimation task. Hence, only the point-cloud encoder is swapped during the experiments. Additionally, we explore whether incorporating point-cloud reconstruction as a secondary task can further enhance the model's performance in mass estimation. For our RGB encoder, we use a DenseNet-121 \cite{b20}, mainly for its small size and good performance on various computer vision benchmarks. For the point-cloud encoders, we test three different encoder variations. The first processes point independently \cite{b16}, the second considers local context and fuses multiple scales \cite{b17}, and the third employs a transformer-based architecture with local context \cite{b21}. Finally, we use a FoldingNet \cite{b22} decoder to reconstruct a coarse point-cloud when performing the reconstruction task.

\subsection{PointNet}

\begin{figure}[t]
    \centering
    \setlength{\fboxrule}{0pt}
    \framebox{\includegraphics[width=0.48\textwidth]{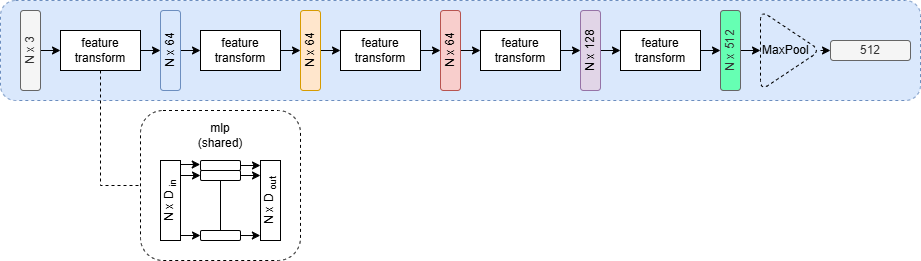}}
    \caption{The PointNet architecture.}
    \label{fig:PointNet}
\end{figure}

Inspired by the PointNet \cite{b16} architecture, this encoder comprises a series of transform blocks that apply a linear transformation to each point separately, as shown in Figure \ref{fig:PointNet}. Inside the same block, the linear transformation is the same for all points; i.e. they share the same parameters. This ensures the network remains invariant to the order of the input points in the point-cloud since point-clouds are naturally unordered data structures. The model progressively transforms the input into higher-dimensional feature spaces until it reaches the maximum feature size of 512 features, at which point a max pool operation is performed to the input, resulting in a 1D vector that works as an embedding.

\subsection{DGCNN}

\begin{figure}[t]
    \centering
    \setlength{\fboxrule}{0pt}
    \framebox{\includegraphics[width=0.48\textwidth]{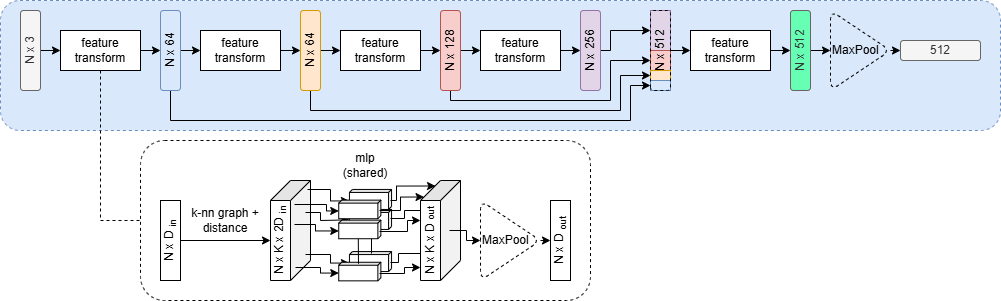}}
    \caption{The DGCNN architecture.}
    \label{fig:DGCNN}
\end{figure}

In our second architecture, inspired by the DGCNN (Dynamic Graph CNN) \cite{b17} paper, illustrated in Figure \ref{fig:DGCNN}, is an evolution of the PointNet model that introduces the concept of dynamically computing local geometric structures via k-nearest-neighbour (k-NN) graphs based on the Euclidean distance. Its feature transformation is very similar to PointNet, but each point \(x_{i}\) is stacked k times, and then, for each neighbour \(x_{j}\), their difference is appended \([x_{i} - x_{j}, x_{i}]\). Then, in a similar fashion to PointNet, a shared linear layer works independently on every point; however, now there are \(N \times k\) points. A MaxPool operation is used to aggregate the linear layer's output. Finally, DGCNN also makes use of residual connections from the outputs of first layers to the last, allowing the network to retain both fine-grained local details and higher-level abstract features.

\subsection{PointTransformer}

\begin{figure}[thpb]
    \centering
    \setlength{\fboxrule}{0pt}
    \framebox{\includegraphics[width=0.48\textwidth]{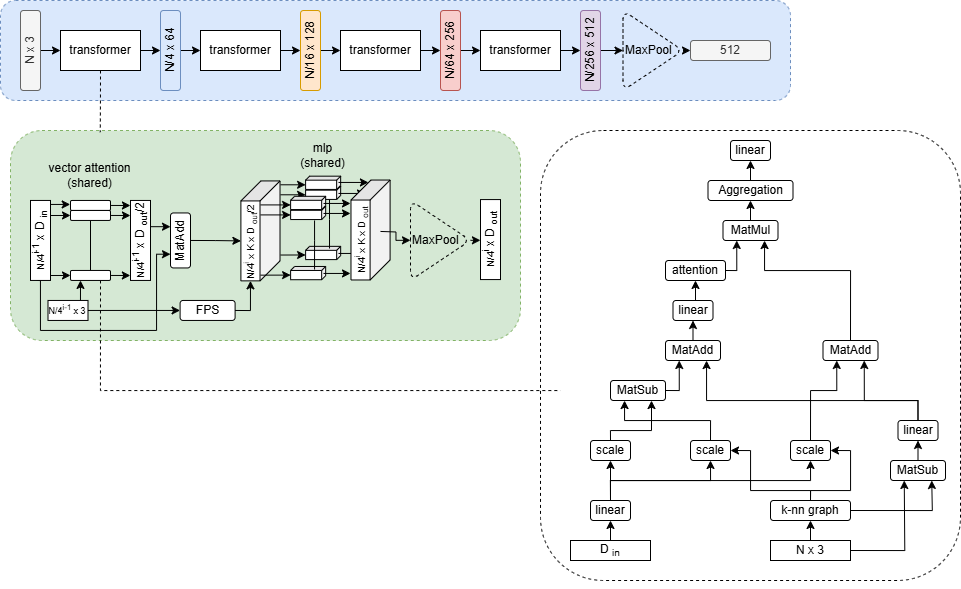}}
    \caption{The architecture of the PointTransformer. A detailed close-up of the transformer block is shown on the bottom left, and to its right, the explicit vector attention mechanism is depicted.}
    \label{fig:PointTransformer}
\end{figure}

Finally, the PointTransformer \cite{b21}, shown in Figure \ref{fig:PointTransformer}, employs a vector attention mechanism to capture descriptions for each point and its local neighbourhood. Vector attention distinguishes itself from traditional attention by considering the k-NN of each point instead of just the point itself. This vector attention is applied individually to each point, with shared weights. In contrast to the other two previous models discussed, PointTransformer also progressively downsamples the point-cloud by a quarter at each transformation block, resulting in fewer points with more descriptive features at each step. Finally, similarly to the previous two models, a MaxPool layer is applied to aggregate the features of the local neighbour into a single feature.

\subsection{DenseNet}

DenseNet is a CNN with dense connectivity. It is divided into sets of dense blocks, in which each layer is connected to every other layer in a feed-forward manner. Additionally, DenseNet uses bottleneck layers (typically 1x1 convolutions) to reduce the number of input feature maps, significantly improving computational efficiency without sacrificing performance.

\subsection{Density Decoder}

The density decoder is a Fully Connected network acting as a regressor for density. Since typical density values are large (in the range of 500 Kg/m³ and above), we fitted a function to the density distribution in the \textit{image2mass} dataset and used it as the activation function for the final layer \cite{b9}. This custom activation introduces an inductive bias, guiding the network toward realistic density estimates. While this bias could potentially limit generalizability to datasets with different density distributions, we argue that the large and diverse nature of the \textit{image2mass} dataset provides a close approximation of real-world object densities when considering similar household settings. Furthermore, this approach reduces implausible predictions and stabilizes training as the network learns to fit realistic density values early on.

\subsection{Volume Decoder}

The volume decoder shares the same structure with the density decoder, except for the fact it has a ReLU has the activation function of the last layer. This results in volumes only having positive values. During training, the density decoder activation function will also guide this network since the two are multiplied to get the final output.

\subsection{FoldingNet}

The folding net serves as a point-cloud reconstruction decoder. It deforms a grid of points by sequentially applying a series of Multi-Layer Perceptrons (MLPs) to each point separately, which translates the point to a new 3D position. The MLPs use the latent vector predicted by the encoders as additional context for the deformation. Furthermore, the same MLPs are used for all points. The output of the FoldingNet is a coarse reconstruction of the input point-cloud.

\subsection{Mass output}

The model computes mass as the product of the outputs from the density and volume decoders. Since the density decoder typically outputs values several orders of magnitude larger than the volume decoder, we introduce a balancing constant \(b\) to scale the two outputs appropriately. By selecting \(b\) such that \(mass = (density \cdot b) \cdot (volume / b)\), and ensuring that \((density \cdot b)\) and \((volume / b)\) are of similar magnitude, the model learns more effectively. Drawing from the approach used in \cite{b9}, we found that a constant \(b \approx 16.5\) yields good results.

\subsection{Metrics}

We use several scale-invariant metrics to evaluate the model’s mass predictions, accounting for the wide variation in object masses. All metrics presented are symmetrically for both over- and under-estimating.

\begin{itemize}
  \item \textbf{Absolute Log Difference Error (ALDE):} \( ALDE = | \ln(y) - \ln(\widehat{y}) | \). This metric measures the difference between the logarithms of the true mass (\(y\)) and the predicted mass (\(\widehat{y}\)). 

  \item \textbf{Absolute Percentage Error (APE):} \( APE = \left| \frac{y - \widehat{y}}{y} \right| \). APE captures the relative error between the true and predicted masses.
  

  \item \textbf{Minimum Ratio Error (MnRE):} \( MnRE = \min \left( \left| \frac{\widehat{y}}{y} \right|, \left| \frac{y}{\widehat{y}} \right| \right) \). This metric measures how much the prediction deviates from the true value in terms of multiplicative factors, making it easy to interpret errors in practical terms. The higher the MnRE, the closer the prediction is to the true value.

  \item \textbf{q-metric (off by a factor of 2):} \( q = \text{Percentage of predictions where } MnRE < 0.5 \). This metric measures the percentage of predictions that are off by a factor of 2 or more. It provides some insight into the consistency and reliability of the model. 
\end{itemize}

\subsection{Losses}

We implement two distinct loss functions depending on whether we are training solely for mass estimation or also for point-cloud reconstruction.

For mass estimation, we use the ALDE loss. We believe ALDE is preferred over metrics like APE because it is less sensitive to extreme values, which can disproportionately penalize errors in smaller mass predictions.

For the task of point-cloud reconstruction, we use the Chamfer Distance, which measures the average distance between points in the predicted and ground-truth point-clouds. This metric is defined as:

\[
\text{CD} = \frac{1}{|P|} \sum_{p \in P} \min_{p' \in P'} \| p - p' \|^2 + \frac{1}{|P'|} \sum_{p' \in P'} \min_{p \in P} \| p' - p \|^2 ,
\]

where \(P\) and \(P'\) are the sets of points in the predicted and ground-truth point-clouds, respectively.

When training the model for both mass estimation and point-cloud reconstruction, we combine these two loss functions into a single objective:

\[
\text{Total Loss} = \text{ALDE} + \lambda \cdot \text{Chamfer Distance} .
\]

Here, \(\lambda\) is a constant factor used to balance the relative contributions of the ALDE and Chamfer Distance components in the total loss function. From empirical experiments, we found that setting \(\lambda \approx 1 \) provides a good balance between mass estimation accuracy and point-cloud reconstruction quality.

\section{RESULTS AND DISCUSSION}

\subsection{Depth Prediction}


We train our GLPDepth model on our synthetic dataset constructed from ShapeNetSem. This dataset includes 8,948 distinct 3D models, each rendered from 14 different viewpoints. To ensure proper evaluation, we divided the dataset into training and testing sets using a 90-10 split, yielding 8,053 models for training and 895 3D models for testing. This corresponds to 112,742 training samples and 12,530 test samples, with no overlap between the 3D models in the train and test sets. This separation ensures that the objects seen during training are not present in the test set.

We use several metrics to evaluate GLPDepth dense depth map reconstructions:

\begin{itemize}
  \item \textbf{Mean Absolute Percentage Error:}
  
  \( mAPE =  \sum_{i=1}^{N} \left| \frac{y - \widehat{y}}{y} \right| . \)
  
  \item \textbf{Mean Square Percentage Error:}
  
  \( mSPE =  \sum_{i=1}^{N} \left( \frac{y - \widehat{y}}{y} \right)^{2} . \)
  
  \item \textbf{Root Mean Square Error:}
  
  \( RMSE = \sqrt{ \frac{1}{N} \sum_{i=1}^{N} (y_i - \widehat{y}_i)^{2} } . \)
  
  \item \textbf{Root Mean Square Logarithmic Error:}
  
  \( RMSE_{log} = \sqrt{ \frac{1}{N} \sum_{i=1}^{N} \left( \log(y_i + 1) - \log(\widehat{y}_i + 1) \right)^{2} } . \)
  
  \item \textbf{Log10 Error:}
  
  \( Log10 = \frac{1}{N} \sum_{i=1}^{N} \left| \log_{10}(y_i) - \log_{10}(\widehat{y}_i) \right| . \)
  
  \item \textbf{Scale Invariant Logarithmic Error:} 
  
  \( SILog = \frac{1}{N} \sum_{i=1}^{N} \left( \log(y_i) - \log(\widehat{y}_i) \right)^{2} - \frac{1}{N^{2}} \left( \sum_{i=1}^{N} \log(y_i) - \log(\widehat{y}_i) \right)^{2} . \)
\end{itemize}

\begin{itemize}
  \item \textbf{Absolute Log Difference Error:}
  
  \( ALDE = | \ln(y) - \ln(\widehat{y}) | . \)
  
  \item \textbf{Absolute Percentage Error:}
  
  \( APE = \left| \frac{y - \widehat{y}}{y} \right| . \)
  
  \item \textbf{Minimum Ratio Error:}
  
  \( MnRE = \min \left( \left| \frac{\widehat{y}}{y} \right|, \left| \frac{y}{\widehat{y}} \right| \right) . \)
  
  \item \textbf{q-metric (off by a factor of 2):}
  
  \( q = \text{Percentage of predictions where } MnRE < 0.5 . \)

  \item \textbf{Chamfer Distance:}
  
  \( \text{CD} = \frac{1}{|P|} \sum_{p \in P} \min_{p' \in P'} \| p - p' \|^2 + \frac{1}{|P'|} \sum_{p' \in P'} \min_{p \in P} \| p' - p \|^2 . \)
\end{itemize}

We use the SILog loss function for training, and fine-tuned two GLPDepth NYU-v2 \cite{b23} checkpoints—one without depth normalization and another with depth normalization. Both models were fine-tuned for 5 epochs. The results of these fine-tunings are presented in Table \ref{fig:table1}.

\begin{figure}[thpb]
    \centering
    \setlength{\fboxrule}{0pt}
    \framebox{\includegraphics[width=0.48\textwidth]{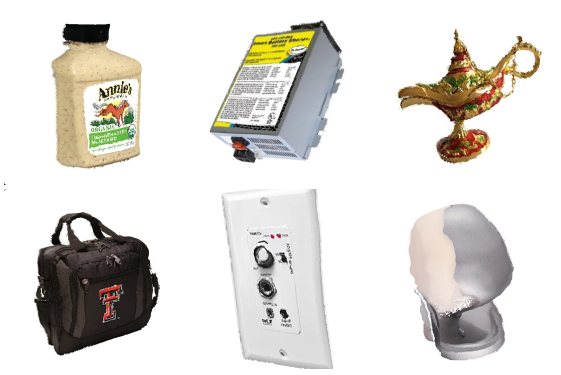}}
    \caption{High definition point-clouds generated from depth predictions of the \textit{image2mass} dataset using the Glpdepth model.}
    \label{fig:GLPDepth}
\end{figure}

\begin{table}[h]
    \centering
    \caption{Glpdepth Performance on Synthetic Test Dataset (12,530 examples)}
    \scalebox{1}{%
    \begin{tabular}{lcccccc}
        \toprule
        \textbf{Model} & \textbf{mAPE ($\downarrow$)} & \textbf{mSPE ($\downarrow$)} \\
        \midrule
        GLPDepth fine-tune     & 0.2028 & 0.1682 \\
        GLPDepth fine-tune + normalization & \textbf{0.0793} & \textbf{0.0181} \\
        \midrule
        GLPDepth Best (20 epoch) & \textbf{0.0718} & \textbf{0.0155} \\
        \midrule
        \textbf{Model} & \textbf{RMSE ($\downarrow$)} & \textbf{RMSE Log ($\downarrow$)} \\
        \midrule
        GLPDepth fine-tune      & 0.4718 & 0.2119 \\
        GLPDepth fine-tune + normalization & \textbf{0.1175} & \textbf{0.0813} \\
        \midrule
        GLPDepth Best (20 epoch) & \textbf{0.1056} & \textbf{0.0741} \\
        \midrule
        \textbf{Model} & \textbf{Log$_{10}$ ($\downarrow$)} & \textbf{SILog ($\downarrow$)} \\
        \midrule
        GLPDepth fine-tune      & 0.0880 & 0.1582 \\
        GLPDepth fine-tune + normalization & \textbf{0.0329} & \textbf{0.0629} \\
        \midrule
        GLPDepth Best (20 epoch) & \textbf{0.0302} & \textbf{0.0567} \\
        \bottomrule
    \end{tabular}
    \label{fig:table1}
    }
\end{table}

The results demonstrate that the normalization procedure significantly enhanced the model’s performance, as evidenced by superior results across all metrics by a wide margin. This intuitively makes sense since it leveraged the 3D structure of the data to resolve the scale ambiguity of the problem. We further fine-tuned the GLPDepth model with normalization for an additional 15 epochs (GLPDepth Best), resulting in even better performance. Although the improvements were evident across all metrics, the difference between the 20-epoch model and the 5-epoch version was less pronounced.

When applied to the \textit{image2mass} dataset, the model's depth predictions, as shown in Figure \ref{fig:GLPDepth}, were qualitatively impressive, providing realistic 3D projections of many objects. However, we also noticed the model struggled with objects with highlights, likely due to the synthetic dataset’s lack of highlight simulation and reflective materials, which probably impacted generalization in such cases. 

\subsection{Mass estimation}

We trained five variations of our original architecture and compared them against the benchmarks from the \textit{image2mass} paper:

\begin{itemize}
  \item \textbf{DenseNet:} A baseline setup where no depth image is provided to the model, and no point-cloud encoder is used.

  \item \textbf{DenseNet + PointNet:} A setup using FoldingNet as the point-cloud encoder.

  \item \textbf{DenseNet + DGCNN:} A setup using DGCNN as the point-cloud encoder.

  \item \textbf{DenseNet + PointTransformer:} A setup using PointTransformer as the point-cloud encoder.

  \item \textbf{DenseNet + PointNet + FoldingNet:} A setup that tests whether simultaneously learning point-cloud reconstruction helps improve the mass prediction capabilities of the model.
\end{itemize}

All models were trained on the same data split as the original \textit{image2mass} experiments, comprising 141,950 training samples and 924 test samples. Importantly, the augmented \textit{image2mass} dataset used in our models contains identical image and mass labels as the original but includes additional point-cloud data.

In the DenseNet setupt, to compensate for the absence of a point-cloud encoder, we increase the size of the DenseNet output latent vector to 1024. In the DenseNet + PointNet + FoldingNet configuration, our synthetic \textit{ShapeNetSem} dataset was used in conjunction with the augmented \textit{image2mass} dataset. Since the \textit{image2mass} dataset lacks 3D models necessary for point-cloud reconstruction, we employed a dynamic batching strategy. Before each training epoch, batches were prepared from both datasets and randomly selected during training. For batches from \textit{ShapeNetSem}, the loss function included both Chamfer Distance and ALDE; for batches from the \textit{image2mass} dataset, only ALDE was used.

To ensure robustness during training, several augmentation techniques were applied. Images were randomly flipped horizontally, vertically, or both, with corresponding transformations applied to the point-clouds. For models with point-cloud encoders, depth images were converted into sparse point-clouds of 1024 points. If the point-cloud had more than 1024 points, random sampling was employed to reduce the number of points to 1024. Conversely, if fewer points were available, additional points were placed at the origin to reach 1024. The point-clouds were centered by subtracting the mean coordinates, computed either after downsampling, or at the beginning of the pre-processing for smaller point-clouds.

\begin{table}[h!]
    \centering
    \caption{Model Performance on the \textit{image2mass} Test Set (924 items)}
    \scalebox{0.83}{%
    \begin{tabular}{lcccccc}
        \toprule
        \textbf{Model} & \textbf{ALDE ($\downarrow$)} & \textbf{APE ($\downarrow$)} & \textbf{MnRE ($\uparrow$)} & \textbf{q ($\uparrow$)} \\
        \midrule
        (i2m) Xception k-NN      & 0.914 & 2.762 & 0.504 & 0.488 \\
        (i2m) Xception k-NN (SIM)& 0.677 & 1.14 & 0.570 & 0.584 \\
        (i2m) Pure CNN           & 0.550 & 0.711 & 0.634 & 0.696 \\
        (i2m) CNN with ALDE      & 0.592 & 0.849 & 0.616 & 0.672 \\
        (i2m) No Geometry        & 0.490 & 0.613 & 0.657 & 0.746 \\
        (i2m) Shape-aware        & 0.470 & 0.651 & 0.672 & 0.767 \\
        (ours) DenseNet                 & 0.623 & 0.879 & 0.612 & 0.663 \\
        (ours) DenseNet + PointTransformer   &  0.477 & 0.624 & 0.671 & 0.757 \\
        (ours) DenseNet + DGCNN              & 0.448 & 0.607 & 0.682 & 0.783 \\
        (ours) DenseNet + PointNet           & \textbf{0.430}  & \textbf{0.522} & \textbf{0.698} & \textbf{0.800} \\
        (ours) DenseNet + PointNet + FoldingNet  & 0.435 & 0.546 & 0.695 & \textbf{0.805} \\
        \bottomrule
    \end{tabular}
    \label{fig:table2}
    }
\end{table}

Table \ref{fig:table2} shows the performance of our models on the \textit{image2mass} dataset. Models starting with (i2m) refer to the \textit{image2mass} benchmarks \cite{b9}. Our PointNet and DGCNN models outperform the original \textit{image2mass} architecture across all metrics. Among the point-cloud encoders tested, PointNet achieved the best performance, which is somewhat surprising, given that it's the simplest of the three models and it doesn't consider local connectivity like the others do. Additionally, the DenseNet-only baseline performed considerably worse than all other proposed models, underscoring the critical role that depth plays in mass estimation.

We also tested PointNet with a FoldingNet, to averiguate whether mass estimation and point-cloud reconstruction could aid each other. We found that both PointNet and PointNet + FoldingNet performed very similarly in mass estimation, which might suggest that point-cloud reconstruction and mass prediction are sufficiently different tasks that improvements in one do not directly translate to improvements in the other. However, this result might also be partly caused by the synthetic dataset being the only one with a complete point-cloud for reconstruction.


\section{CONCLUSION}

Mass estimation using purely visual methods remains an underexplored area in robotics, with current approaches relying on images and bounding boxes to estimate object mass. However, these methods are often impractical due to their slow performance, requiring multiple views of the object. To address this challenge, we augmented an existing RGB-mass dataset with depth data, and reformulated the task as predicting mass from both depth and RGB inputs. We developed a network that integrates sparse point clouds and RGB images to estimate object mass and tested several point cloud encoders, including PointNet, DGCNN, and PointTransformer. Our model outperformed existing models on the same dataset, with PointNet yielding the best results, unexpectedly surpassing the more complex architectures.

Additionally, we explored whether learning to reconstruct a complete point cloud could improve mass estimation, though the results were inconclusive. As a next step, we aim to deploy this model in practical robotic applications, investigating how planning and control algorithms could leverage mass predictions to enhance performance in real-world tasks such as grasping and manipulation.

\section*{Acknowledgements}
This work was supported by LARSyS FCT funding (DOI: 10.54499/LA/P/0083/2020, 10.54499/UIDP/50009/2020, and 10.54499/UIDB/50009/2020), the H2020 FET-Open project RePAIR under EU grant agreement 964854, by the Lisbon Ellis Unit, and FCT project 2024.07248.IACDC.

\end{document}